%% file: 00-NNIR-2016-main.tex
\documentclass{sig-alternate-05-2015}
\usepackage{color}
\usepackage{balance}
\usepackage{soul}

\permission{Permission to make digital or hard copies of part or all of this work for personal or classroom use is granted without fee provided that copies are not made or distributed for profit or commercial advantage and that copies bear this notice and the full citation on the first page. Copyrights for third-party components of this work must be honored. For all other uses, contact the owner/author(s).}

\begin{document}




\conferenceinfo{\hspace{-0.5mm}Neu-IR'16 SIGIR Workshop on Neural Information Retrieval,}{July 21, 2016, Pisa, Italy\\\\ \copyright\,$ 2016$ Copyright is held by the owner/author(s).}


%

\title{Emulating Human Conversations using Convolutional Neural Network-based IR}
%
%
%
%
%

\numberofauthors{3} 
%
\author{
%
%
\alignauthor
Abhay Prakash\\
      \affaddr{Microsoft, India}\\
     \email{abprak@microsoft.com}
\alignauthor
Chris Brockett\\
      \affaddr{Microsoft Research,}\\
    \affaddr{ Redmond, WA, USA}\\
    \email{chrisbkt@microsoft.com}
\alignauthor  
Puneet Agrawal\\
      \affaddr{Microsoft, India}\\
    \email{punagr@microsoft.com}
}

\maketitle
\begin{abstract}
Conversational agents (``bots'') are beginning to be widely used in conversational interfaces. To design a system that is capable of emulating human-like interactions, a conversational layer that can serve as a fabric for chat-like interaction with the agent is needed. In this paper, we introduce a model that employs Information Retrieval by utilizing convolutional deep structured semantic neural network-based features in the ranker to present human-like responses in ongoing conversation with a user. In conversations, accounting for context is critical to the retrieval model; we show that our context-sensitive approach using a Convolutional Deep Structured Semantic Model (cDSSM) with character trigrams significantly outperforms several conventional baselines in terms of the relevance of responses retrieved.
\end{abstract}

%
%


%
%

%
%


\category{H.3.3}{Information Storage And Retrieval}{Information Search and Retrieval}
\category{I.2.7}{Artificial Intelligence}{Natural Language Processing}


\keywords{Chat bot; Deep learning; Structured Semantics; Conversational agent; Convolutional Networks; Twitter data}

\section{Introduction}
\label{sec:intro}
\input{01-intro.tex}

\section{Related Work}
\label{sec:related}
\input{02-related.tex}

\section{Our Approach}
\label{sec:approach}
\input{03-00-approach.tex}

\subsection{Data for Index and Ranker}
\label{ssec:data}
\input{03-01-data.tex}

\subsection{Retrieving Candidates}
\label{ssec:retrieval}
\input{03-02-retrieval.tex}

\subsection{Convolutional DSSM}
\label{ssec:dssm}
\input{03-03-DSSM.tex}

\subsection{Learning to Rank in Multi-Turn Chat}
\label{ssec:learn2rank}
\input{03-04-learn2rank.tex}

\section{Evaluation}
\label{ssec:eval}
\input{04-eval.tex}


\section{Conclusions}
\label{sec:conclusions}
\input{06-conclusions.tex}

\section{Acknowledgments}
We thank Bill Dolan, Michel Galley, Jianfeng Gao, Manoj K Chinnkotla, Chakrapani Ravi Kiran, Puneet Garg and Radhakrishnan Srikanth for helpful discussions and comments. We would also like to thank the two anonymous reviewers for their comments. 

%
\balance
\bibliographystyle{abbrv}
\bibliography{sigproc}

\end{document}

%% file: 01-intro.tex
Recent research in neural network-based Question Answering (QA), has focused primarily on providing the most relevant answer to a given query \cite{qiu2015convolutional}\cite{severyn2015learning}\cite{shang2015neural}\cite{shang2016shorttext}\cite{yin2015neural}\cite{yu2014deep}. 
However, as businesses move towards building conversational interfaces (for example, Facebook M, Cortana, Siri, and Orat.ai) where humans and systems work collaboratively to achieve their goals, it becomes important to model the context surrounding the conversation, since the system is not only responding to a question, but needs to answer in the context of the history of the exchange.

Context is crucial, because conversational language does not always make for good query terms. Figure \ref{example_conversation} presents an example of an interaction between a conversational agent (``bot'') and user that resembles a dialog between friends. The bot responds, on the basis of the previous history of the conversation, \textit{mine is 25 June}. It will be noted, however, that the user's messages in this conversation are casually colloquial and, taken in isolation, vaguely worded (\textit{cool. yours on?}). Such queries are not easily addressable in traditional QA systems, which might need to perform complex anaphora resolution to determine that the query \textit{yours on?} relates to the phrase \textit{My bday month}.  

\begin{figure}[t]
\begin{tabular}{ll}
\\
    \textbf{User:}&\textit{June starts. My bday month. aww...}\\
    \textbf{Bot:}&\textit{for me too}\\
    \textbf{User:}&\textit{cool. yours on?}\\
    \textbf{Bot:}&\textit{mine is 25 June}
\end{tabular}
\caption{\small Example conversation between user and bot.}
\label{example_conversation}
\end{figure}

In this paper, we present a deep neural network-based approach to implementing a conversational agent that will engage with users in a friendly, engaging, conversational fashion, drawing on a database of Twitter conversations. We model the task of providing a response as an Information Retrieval problem, i.e., for a given query issued by the user, the bot must retrieve the best candidate from a conversational corpus to output as response. Below, we use the following notation: we will refer to user message as M, agent response as R, and the previous history of conversation as C (context). We use features derived from a Convolutional Deep Structured Semantic Model (cDSSM) \cite{shen2014learning} to predict the output R given (M, C) on a corpus of Twitter conversations and demonstrate that use of cDSSM-based contextual features can boost the 1-best precision of retrieved responses over several alternatives.

%% file: 02-related.tex
\begin{figure}[!t]
\centering
\includegraphics[scale = 0.7]{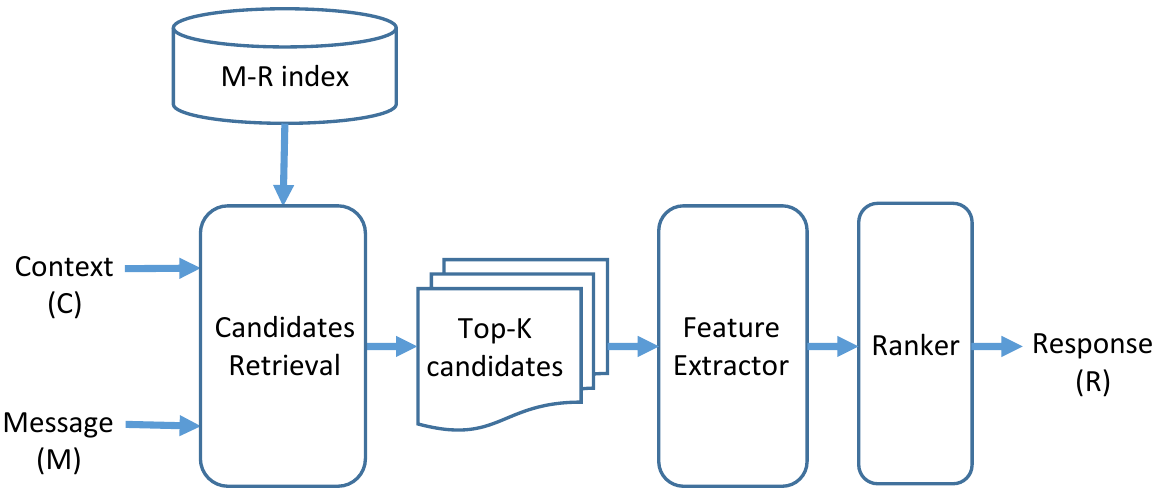}
\caption{\small Our system is similar to a traditional IR system. We focus on extracting relevant features using deep learning.}
\label{Architecture}
\end{figure}

A growing body of research has emerged in chat response generation, either using machine translation techniques \cite{ritter2011data} or deep neural networks, including sequence-to-sequence modeling \cite{shang2015neural}\cite{sordoni2015neural}\cite{vinyals2015neural}. These models, however, can be cumbersome to train, and  may suffer a tendency towards blandness of output \cite{li2015diversity}.

The system we explore here, by contrast, uses deep learning methods to retrieve the best response from a database of candidates. These candidates come from real human conversations and hence may be more vivid and less consensus-driven. In this respect, the system is close in spirit to the work of \cite{swanson2008say} on interactive storytelling where for a given message (\textit{user's sentence}) the system retrieves the best matching message (\textit{a sentence in story corpus}) and returns with the corresponding response (\textit{following sentence in story corpus}), using TF-IDF-based scoring. 

The present work also bears a familial resemblance to \cite{jafarpour2010filter} and the IR baseline in \cite{sordoni2015neural}, where Twitter conversation pairs (each consisting of a tweet and its response) are used as data. Like our system, for a given query, those authors retrieve the best matching tweets to form a candidate set by taking their corresponding replies and rank them to display an appropriate response.
\cite{jafarpour2010filter} does not take into account context and does not deploy deep learning.  
\cite{sordoni2015neural} describe a neural language generation model (DCGM-II + CMM) that incorporates context to rank retrieval candidates. We use cDSSM, which is computationally efficient and relies on implicit semantic structure to derive features used in the ranker (described in Section \ref{ssec:learn2rank}). 

Deep learning techniques have recently been applied in direct community question-answering in \cite{qiu2015convolutional}\cite{severyn2015learning}\cite{yu2014deep}, and also in \cite{yih2014semantic} who applied cDSSM. These approaches seek to return the answer from a knowledge base in triplet format. 
Similarly, \cite{severyn2015learning} used a convolutional neural network approach to rank short text pairs when answering questions.
Other recent neural retrieval models include a multilayer perceptron classifier \cite{sugiyama2016} and three-layered neural networks \cite{Denawa2016}\cite{liu2016} in the NTCIR-12 short text conversation tasks. 
In these models, however, there is no dependency on the history of conversation to determine the user's intent.

%% file: 03-00-approach.tex
We model the task of providing appropriate chat responses as an Information Retrieval problem, where for a given user message M and context C, the system retrieves and ranks the candidates by relevance and outputs one of the highest scoring responses. Offline, we create an index of paired tweets and their responses and index these using Lucene\footnote{http://lucene.apache.org/}. 
At runtime, the best response is chosen in a three step process. First, we use TF-IDF-based fetch to generate a candidate set appropriate to M and C. Then we extract features using a convolutional deep structured semantic network. Finally, a ranker is trained on 3-turn twitter conversations using these features to select response R from the candidate set (described in sections \ref{ssec:learn2rank} below). Our setting is similar to any traditional IR system, and our main focus is on extracting relevant features and improving the candidate set selection.
The runtime process is depicted in Figure \ref{Architecture}.

%% file: 03-01-data.tex
\subsubsection{Data preparation for M-R pair Index}
\label{sssec:dataprep4index}

We constructed a dataset of 17.62 million tweet conversational pairs (tweets and their responses), extracted from the Twitter Firehose, covering the four year period from 2012 through 2015. To favor responses reflecting a culturally local persona, we limited the geographical region to a specific time zone. This permitted us to expose more culturally-appropriate responses, for example, the query \textit{what do you like for dinner} triggers the response \textit{bhindi masala} (an Indian curry made with okra) for Indian users and \textit{kaeng lueang (``yellow curry'')} for users in Thailand.

In order to protect privacy and prevent personal information from surfacing in our chat agent's responses, we removed from this dataset, any conversational pairs where the response contained any individual's name, URL or hashtag. Further, we sought to minimize the risk of offending users by removing any pairs in which either M or R contained adult, politically sensitive, or ethnic-religious content, or other potentially offensive or contentious material, such as inappropriate references to violence, crime and illegal substances. We also filtered data to avoid scenarios where an adversarial query would result in retrieval of a candidate with high confidence from index like \textit{Does <celebrity> hate <religion>?} --> \textit{Yes, he does}, where <celebrity> and <religion> stand in for specific entities\footnote{Queries involving such potentially offensive material must be detected and handled in a separate  component that is beyond the scope of this paper.}. 
After filtering, our corpus comprises 9.56 million usable M-R pairs. 

\subsubsection{Training Data for the Ranker}\label{sssec:training_data_ranker}
To train the ranker, we obtained a 2.58 million sample of 3-turn tweet conversations from the Twitter Firehose, in the same manner as described in Section \ref{sssec:dataprep4index} above. Each conversation comprises an exchange between two Twitter users, alternating over 3 turns.

%% file: 03-02-retrieval.tex
The goal of the retrieval step is to fetch, with high recall, relevant documents present in the index for the given M and C. As noted above, we have built an index of messages and responses. Given M, this index is used to retrieve best matching messages and their corresponding responses that constitute the candidate set. 
From a retrieval perspective, however, many messages in conversational scenarios are not self-sufficient as queries to return a good candidate set, but are dependent on expressions in the previous context e.g. \textit{why?}, \text{go ahead}, and \textit{please}. For shorter queries such as these, we fetch documents after appending context to query. This solution, though simple, works pragmatically and does not form the main focus of this paper.

%% file: 03-03-DSSM.tex
\begin{figure}[!t]
\centering
\vspace{0.1 cm}
\includegraphics[scale = 0.76]{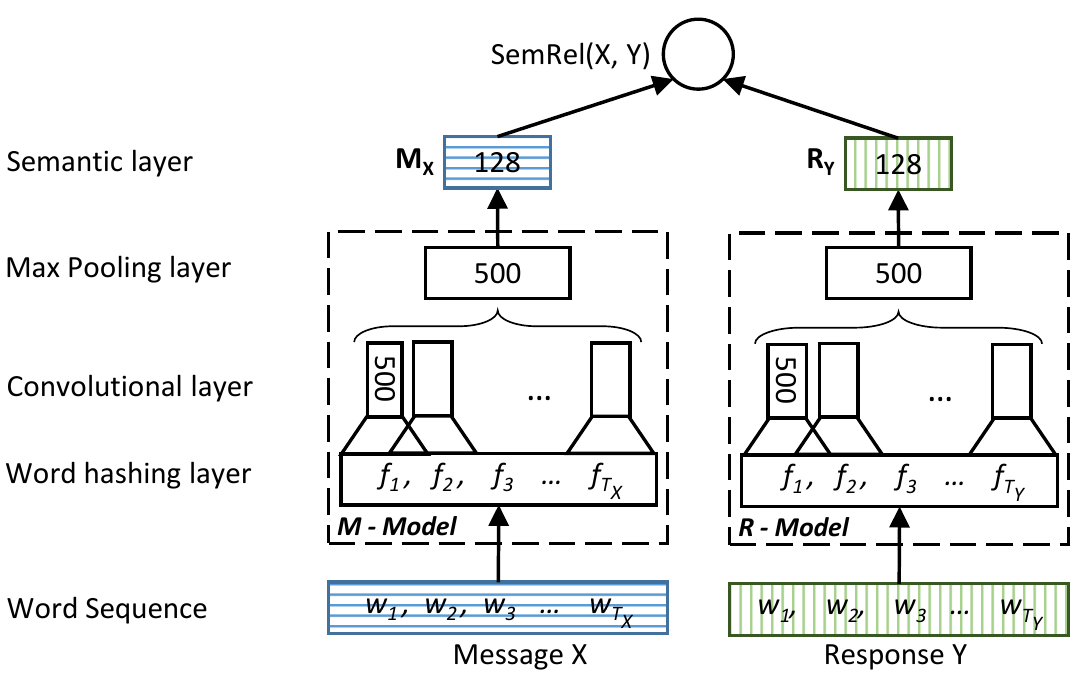}
\vspace{0.05 cm}
\caption{\small cDSSM uses a pair of models to vectorize message and response to their respective low-dimensional vectors.}
\vspace{0.05 cm}
\label{fig:cdssm}
\end{figure}

To extract features for our ranker, we trained a Convolutional Deep Structured Semantic Model (cDSSM) \cite{shen2014learning}. 
cDSSM finds semantic relevance between text query and document, after being trained on click-through data to  maximize conditional likelihood of the clicked documents for the given query. 
We adapted this formulation to our purpose by utilizing M-R pairs from Twitter (described in \ref{sssec:dataprep4index} above) as training data, where the user message is treated as query and response as clicked document. 
The cDSSM model thus learns the semantic relevance between a message and its response. Note that semantic relevance indicates, for a given M-R pair, how relevant R is for M, and is distinct from semantic similarity. We chose cDSSM over DSSM\footnote{Initial experimentation with DSSM failed to yield useful results.} \cite{huang2013learning} in order to incorporate structural (word position) features in both message and response. To prepare the character trigram set \cite{huang2013learning}, we took the most frequent 5k character trigrams found in the Twitter dataset.

Figure \ref{fig:cdssm} shows two models, a Message model (M-Model) and a Response model (R-Model), that have been obtained after training on M-R pairs. When a text X is forward propagated through either of the models, it is vectorized in the space defined by that model. We use the notation $\mathbf{M_{X}}$ for the vector when X is vectorized using the M-model and $\mathbf{R_{X}}$ when it is vectorized using the R-Model. We define two scoring functions using the two models as follows:
\newline
\newline
\textbf{Semantic Relevance Score(X,Y):} We denote this score as \textit{SemRel(X, Y)}. This is the confidence of semantic relevance of Y as a response to X, calculated as follows: 
\begin{equation*}
SemRel(X,Y) = cosine(\mathbf{M_{X}}, \mathbf{R_{Y}}) = \frac{\mathbf{{M_{X}}^{T}\mathbf{R_{Y}}}}{\mathbf{\|M_{X}\|\|R_{Y}\|}}
\end{equation*}
\newline
\textbf{Semantic Similarity Score(X,Y):} This score is denoted as \textit{SemSim(X, Y)}. It indicates semantic similarity of two texts X and Y in the space defined by the model used. To compute semantic similarity of two responses X and Y, both X and Y are vectorized using the R-Model, using the following formula:
\begin{equation*}
SemSim(X,Y) = cosine(\mathbf{R_{X}}, \mathbf{R_{Y}}) = \frac{\mathbf{R_{X}}^{T}\mathbf{R_{Y}}}{\|\mathbf{R_{X}}\|\|\mathbf{R_{Y}}\|}
\end{equation*}

The semantic similarity of two messages could also be computed using the M-model in an analogous manner, though we did not use this approach here. We use the above two scoring functions to generate features as described in Section \ref{ssec:learn2rank} below. 

%% file: 03-04-learn2rank.tex
Since we have modeled the problem as ranking task, we trained an implementation of the MART gradient boosting algorithm \cite{burges2010ranknet} to order the responses in candidate set. Note that as cDSSM is trained on only M-R pairs and thus learns to rank candidates only for a given message, we use it to derive features for the ranker to pick the best contextual response in the ongoing conversation.
\newline
\newline
\textbf{Data format for ranker training:} We used 3-turn Twitter conversations as described in Section \ref{sssec:training_data_ranker}. From each conversation, we prepared three training samples (1 positive and 2 negative), where Turn 1 and 2 were taken as C and M respectively. For positive samples, the original actual response in Turn 3 was used. For the two negative samples, random responses were selected (without replacement) from entire set of Tweets.
\newline
\newline
\textbf{Features for ranker training:} We used the following features to train the ranker, which incorporated context while ranking the candidates:
\newline
\textbf{\textit{i) SemRel(M,R)}} -- this represents the relevance of candidate response R for a given M without considering context. This is calculated from the cDSSM model as described in Section \ref{ssec:dssm}. 
\newline
\textbf{\textit{ii) Context Message Match (CMM)}} -- These are the exact matches between C, M and R (borrowed from \cite{sordoni2015neural}). These features incorporate lexical similarity of R with M and C. As described in \cite{sordoni2015neural}, we calculate the number of [1-4] n-gram matches between C and R, and between M and R. These matches are helpful for examples where for message \textit{i love you}, \textit{i love you too!} is a relevant response.
\newline
\textbf{\textit{iii) SemSim(C,R)}} -- This feature captures semantic similarity between C and R, calculated using cDSSM. It helps pick a response that is semantically closer to the context and hence capture the mood or intent of conversation. We discuss it further in Section \ref{subsec:qualitative} with an example. 
\newline
\textbf{\textit{iv) SemRel(C,M)}} -- This feature captures change of context in the ongoing discussion. The score shows the relevance of M being the response for the C, where C was the previous message. The lower the value of this feature, the more likely it is that the user has changed topic and contextual features need not to be assigned higher importance.
\newline
\newline
We refer to the last two features \textit{(iii) and (iv)} jointly as \textbf{Context Capturing Features (CCF)}. 
At training time, our best system with all defined features scored NDCG@1 = 84.83, NDCG@2 = 92.56 and NDCG@3 = 94.02 on our development evaluation set. 

%% file: 04-eval.tex
\subsection{Setup}

\begin{table*}[!t]
\centering
\includegraphics[scale = 0.93]{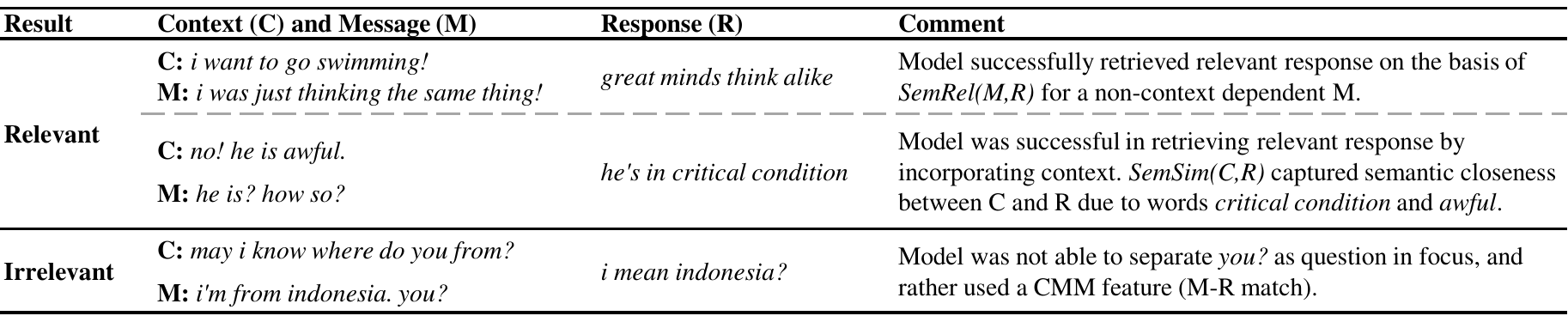}
\vspace{-2mm}
\caption{\small Relevant and irrelevant outputs from the {SemRel(M,R)+CMM+CCF} system, with annotations concerning analysis}
\label{qualitative_analysis}
\vspace{-2mm}
\end{table*}

Since we are primarily interested in the contextual appropriateness of interactions from a human viewpoint, we evaluated two versions of our system, SemRel(M,R) +CMM and SemRel(M,R)+CMM+CCF, using human relevance judgments. 
To this end, we held out 1000 3-turn conversations from our Twitter dataset. 
We took first 2 turns of these 1000 conversations and retrieved the third turn using our system. The retrieved response was shown to human judges and labelled for relevance, using a crowdsourcing platform.
In each case, we employed 5 human judges, who were shown the previous 2 turns by way of context along with the extracted response, and then asked to make a binary decision [0,1] whether or not the response was relevant to the conversational context. 
Other characteristics that might be appropriate in a human-agent interaction, such as interestingness and friendliness were not evaluated at this time. Also, since this is a retrieval-based system, we did not evaluate for grammatical correctness. 

We compared two models against 3 baselines: IR$_{status}$(as in \cite{ritter2011data}), IR$_{status}$+CMM$_{ 10 feat.}$, and DCGM-II+CMM$_{ 10 feat.}$ (as proposed in \cite{sordoni2015neural}). For the purpose of our task, we believe the last is a strong baseline for a retrieval based system, since \cite{sordoni2015neural} also incorporated context when ranking candidates.
When presenting the outputs to the judges, we randomly interpolated the system outputs to prevent the introduction of bias into the judgments.

\begin{table}[t]
\begin{tabular}{lll}
\hline
\textbf{Technique}&\textbf{Score}&\textbf{95\% CI}\\
\hline
IR$_{status}$ & 3.73 & 3.67 3.79\\
IR$_{status}$+CMM & 3.78 & 3.73 3.85\\
\textbf{DCGM-II+CMM} & \textbf{3.87} & 3.80 3.95\\
\hline
SemRel(M,R)+CMM & 4.15 & 4.10 4.21\\
\textbf{SemRel(M,R)+CMM+CCF} & \textbf{4.23} & 4.17 4.29\\
\hline
\end{tabular}
\caption{\small Mean scores assigned by judges, together with 95\% confidence intervals. The best-performing baseline and experimental systems are shown in bold.}
\label{mean_scores}
\end{table}

\begin{table}[t]
\begin{tabular}{lll}
\hline
\textbf{Technique}&\textbf{P@1}&\textbf{95\% CI}\\
\hline
IR$_{status}$ & 62.4 & 59.4 65.4  \\
IR$_{status}$+CMM & 63.8 & 60.8 67.8 \\
\textbf{DCGM-II+CMM} & \textbf{68.4} & 65.5 71.3\\
\hline
SemRel(M,R)+CMM & 79.3 & 76.8 81.8\\
\textbf{SemRel(M,R)+CMM+CCF} & \textbf{82.0} & 79.6 84.4\\
\hline
\end{tabular}
\caption{\small Precision@1 together with 95\% confidence intervals. The best-performing baseline and experimental systems are shown in bold.}
\label{precision_scores}
\vspace{-2mm}
\end{table}

\subsection{Results}
We report the results of human judgments in two ways. First, we compute scores for each of the 1000 responses on a 5 point scale: we convert the binary decisions into a score between 0 and 5 by summing the judgments, then average over those scores. The results are shown in Table \ref{mean_scores}. Two-tailed t-tests indicate that the difference in means between the top-scoring baseline ({DCGM-II+CMM}) and the two SemRel(M,R) systems is statistically significant (p < 0.0001). The SemRel(M,R)+CMM+CCF appears to be slightly better overall; CCF provides a small but useful boost.  

To assess the precision of retrieval, we also used a more conservative measure. Responses that achieved a supermajority of votes ($\geq 4$) were assigned a score of 1, otherwise they were assigned 0. This we will term \textit{Precision@1}, in other words, precision at rank 1. 
The \textit{Precision@1} results are presented in Table \ref{precision_scores}. 
It is evident from the confidence intervals that using features from deep learning models (i.e., cDSSM) to capture deep structured similarity yields significant improvement over other methods.

\begin{figure}[!t]
\begin{tabular}{ll}
\textbf{User:} & \textit{hi, good evening}\\
\textbf{Bot:} & \textit{hey..good evening..}\\
\textbf{User:} & \textit{how are you}\\
\textbf{Bot:} & \textit{i'm fine and you?}\\
\textbf{User:} & \textit{i am also good, just returned from office}\\
\textbf{Bot:} & \textit{where do you work?}\\
\textbf{User:} & \textit{all that later, i have something to tell you}\\
\textbf{Bot:} & \textit{what?}\\
\textbf{User:} & \textit{i like a girl a lot, and want to ask her out}\\
\textbf{Bot:} & \textit{ask her out anyway}\\
\end{tabular}
\caption{\small A sample user interaction with our system illustrating the human-like nature of exchanges.}
\label{fig:actual_Chat}
\end{figure}

The gain in precision between SemRel(M,R)+CMM and the {DCGM-II+CMM} baseline is particularly significant (p < 0.0001) as measured by the Wilcoxon signed-rank test and McNemar's test for dichotomous variables. The addition of CCF features ({SemRel(M,R)+CMM+CCF}) appears to play a useful role in capturing context and in fine tuning ranking according to context.

\subsection{Qualitative Analysis}
\label{subsec:qualitative}

Table \ref{qualitative_analysis} shows examples of relevant and irrelevant results returned by the {SemRel(M,R)+CMM+CCF} model. The examples are taken from our 1000 conversation evaluation set as described in Section \ref{ssec:eval}. The top example in the relevant results section is an instance where we observe same response with or without CCF features. We also note that there are no n-gram matches in C and R. This implies that \textit{SemRel(M,R)} was alone sufficient to retrieve a relevant response, because M itself is descriptive and not heavily dependent on C. The example also suggests that our model does not overweight contextual features.

The second example in the relevant results, shows that the \textit{SemSim(C,R)} feature helps retrieve a relevant response that is consistent with C. Candidate responses for M (\textit{he is? how so?}) included \textit{who he is ?} and \textit{aw tell me about it!! he is so hot!}, which are relevant to M but irrelevant when C is taken into account.

The irrelevant result shown in Table \ref{qualitative_analysis} illustrates a known limitation of our model. Here, an irrelevant response was retrieved because \textit{you?} was the focus of the message: candidates should have been ranked taking into account \textit{you?} as M and \textit{i am from indonesia.} as C. It would appear that none of the candidates received a high score for SemRel(M,R), while the CMM feature for M-R lexical match was assigned high importance.

Figure \ref{fig:actual_Chat} provides a short sample of chat by a human using our system, illustrating the fluency and human-like nature of exchanges.

%% file: 06-conclusions.tex
We have presented initial results of experiments in the use of convolutional Deep Structured Semantic neural networks (cDSSM) to emulate human conversations using IR techniques. Our approach is not only computationally performant but produces results that are significantly more relevant than baseline IR techniques. In these experiments, we employed a simple feature set and focused primarily on learning from cDSSM to output the best responses in context. In the future, we hope to expand the candidate set by introducing better context understanding and by taking into account user sentiment in order to further improve the quality of the human-agent interaction. Our results suggest that cDSSM is a viable approach to emulate interesting conversational exchanges, especially in cases where conversational data may be relatively limited, for example, regional markets and ``smaller'' or minority languages. 